\pdfoutput=1

\documentclass[11pt,twocolumn]{article}

\usepackage{geometry}
\usepackage{authblk}

\usepackage{times}
\usepackage{latexsym}

\usepackage[T1]{fontenc}

\usepackage[utf8]{inputenc}

\usepackage{microtype}

\usepackage{inconsolata}

\usepackage{booktabs}
\usepackage{subcaption}
\usepackage{amsfonts}
\usepackage{amsmath}
\usepackage{multirow}
\usepackage{spverbatim}
\usepackage{float}
\usepackage{xcolor}

\usepackage{graphicx}
\usepackage{enumitem}

\usepackage{graphicx}

\usepackage{url}
\usepackage{verbatim}
\usepackage{natbib}
\title{Exploring Performance Variations in Finetuned Translators of \\ Ultra-Low Resource Languages: Do Linguistic Differences Matter?}

\author[2]{Isabel Gonçalves}
\author[1]{Paulo Cavalin}
\author[1]{Claudio Pinhanez}
\affil[1]{IBM Research Brazil}
\affil[2]{{PUC-Rio}}

\begin{document}
\maketitle
\begin{abstract}
Finetuning pre-trained language models with small amounts of data is a commonly-used method to create translators for ultra-low resource languages such as endangered Indigenous languages. However, previous works have reported substantially different performances with translators created using similar methodology and data. In this work we systematically explored possible causes of the performance difference, aiming to determine whether it was a product of different cleaning procedures, limitations of the pre-trained models, the size of the base model, or the size of the training dataset, studying both directions of translation. Our studies, using two Brazilian Indigenous languages, related but with significant structural linguistic characteristics, indicated none or very limited influence from those training factors, suggesting differences between languages may play a significant role in the ability to produce translators by fine-tuning pre-trained models.
\end{abstract}

\section{Introduction}

There are over 7,000 languages being used today in the world and about half of them are in danger of disappearing until the end of the century~\citep{moseley2010atlas}, of which about half are Indigenous languages. As the use of Internet spreads through those communities, in particular with the arrival of satellite-based connection, it is becoming imperative to provide the members of those communities with tools to facilitate their use in social media, websites, and other electronic written forums~\citep{bird-2020-decolonising}.

Translators for those endangered languages are one of such tools, since it may help bilingual youngsters and adults to leverage what they know in one language to foster the writing in another. However, the construction of such systems using modern AI technology is hindered by the lack of available data for most of those languages, which are often categorized as \textit{ultra-low resource languages}, what prevent the training from scratch of translators and even, having them part of a multilingual translators such as NLLB~\citep{nllbteam2022}.


A commonly-used approach to tackle the lack-of-data issue is to \textit{finetune} pre-trained language models~\citep{raffel2020,gao-etal-2024-novel}. Although the approach has proved to be successful for many languages~\citep{elmadany-etal-2024-toucan}, the technique, when used with ultra-low amounts of finetuning data, is also prone to problems such as overfitting~\citep{elbayad-etal-2023-fixing} and ``rogue'' storage and retrieval of answers~\citep{cavalin-etal-2024-fixing}. However, little is know whether the technique can be applied successfully to a language independently of its structural linguistic characteristics.

This work was motivated by a recent report~\citep{pinhanez2024harnessing} in which translators to Portuguese of two Brazilian Indigenous languages, \textit{Guarani Mbya} and \textit{Nheengatu}, which share similar origins and vocabulary, were built by finetuning the same high-quality translator, WMT-19~\citep{ng-etal-2019-facebook} with datasets of similar characteristics. However, the performance of the two translators is radically different, with the Guarani Mbya and the Nheengatu translators achieving SacreBLEU scores of about 12\% and 38\%, respectively. The work does not provide good reasons for the difference in performance, except that the performance of the Nheengatu translator improved considerably after the training dataset was manually cleaned.

In this work we investigate systematically possible causes of the performance difference, aiming to determine whether it was a product of different cleaning procedures, limitations of the WMT-19 model, the size of the base model, or the size of the training dataset. We also explore the other direction of the translation, from Portuguese to the Indigenous languages, to see whether the issue was related to text generation problems. 

We started by contacting the authors of~\citep{pinhanez2024harnessing}, from whom we obtained the training data. Unlikely what was done in that work, we employed as the pre-trained model, \textit{NLLB-200}~\citep{nllbteam2022}, used in other efforts to create translators~\citep{DERE2023e01809, haberland-etal-2024-italian,robles2024} and with American Indigenous languages~\citep{degenaro-lupicki-2024-experiments,prieto-etal-2024-translation}. 
After systematically cleaning the datasets provided to us, we first performed a \textit{zeroshot} evaluation of the NLLB-200 model with 600 million parameters, obtaining poor but similar performance for Guarani Mbya and Nheengatu, using the standard chrF metrics. We then finetuned the NLLB model with training datasets of about 3K and 7K pairs, respectively, and obtained much better performance for Nheengatu. The performance difference was not significantly affected when we finetuned with a larger version of NLLB-200 with 3 billion parameters. We then experimented with downsampling the Nheengatu training dataset, obtaining only similarly poor performance to the Guarani Mbya translator at extremely low amounts of data. The bulk of this work is the description and analysis of those experiments and results.




Albeit Guarani Mbya and Nheengatu belong to the same linguistic family (see Section~\ref{bils}), the languages also present morphological differences which may directly influence the results, such as notable differences in word order. The main contribution of this work is the exploration of possible causes for the differences in performance seen in the translators for those two languages, ruling out factors such as the pre-trained model used, the direction of translation, model size, and, to a large extent, the size of the training dataset.  Our results seem to indicate, as discussed in detail throughout the paper, that, given that those other factors do not seem to explain the performance differences, that linguistic characteristics may play a significant role in the feasibility of using ``standard'' finetuning methods ultra-low resource languages.

\section{Related work}
Over the most recent years, we have seen progress in research and applications for Natural Language Processing (NLP) and Machine Translation (MT). Foundation models have emerged as an efficient way to leverage general-purpose pre-training efforts to speed up and improve more-specific downstream applications \citep{bommasanifoundationmodel2021}, in a step that we usually refer as finetuning.

Finetuning is especially important for both low- and extremely-low resource languages, since this is a well-known way to take advantage of related data or languages to improve the lack-of-data issue for downstream applications \citep{raffel2020}. Although the process of finetuning can be conducted with more elaborated procedures, such as the two-step finetuning for low-resource languages \citep{gao-etal-2024-novel}, in this work we limit our focus to the traditional single-stop process, i.e. from pre-trained models to a single finetuning step.

One pre-trained model that has been widely used for this purpose is NLLB, from the No Language Left Behind project \citep{nllbteam2022languageleftbehindscaling}. This model has been pre-trained with 200 languages, in many-to-many directions, and has demonstrated notorious success in many settings \citep{DERE2023e01809,haberland-etal-2024-italian,robles2024,basit-etal-2024-challenges}. 

For Indigenous languages, finetuning a model like NLLB has played an important role in achieving better quality for MT models. Some notable works can be found in the AmericasNLP workshop, such as the finetuning of NLLB with 600M parameters on Spanish to 11 Indigenous languages \citep{degenaro-lupicki-2024-experiments} and for two languages from Colombia \citep{prieto-etal-2024-translation}. What is noticeable, however, is that the success varies from language to language without a direct correlation with training set size \citep{degenaro-lupicki-2024-experiments}.





\section{Guarani Mbya and Nheengatu}\label{bils}

\citet{storto2019} provides a good overview of the history, structure, and characteristics of \emph{Brazilian Indigenous Languages}. Those languages are often classified into 4~main groups, according to geo-linguistisc features \citep{rodrigues1986}: the \emph{Macro-J\^{e}} and the \emph{Tupi} branches, comprising each several language families; the languages which belong to an identified \emph{family}, such as the \emph{Aruak}, \emph{Karib}, \emph{Tukano}, \emph{Nambikw\'ara} families; and languages for which no family has been identified, such as \emph{Tikuna} (the most spoken BIL), \emph{Pankarar\'u}, and \emph{Arik\'en}. According to \citet{diniz2007notas}, only about 90 of the BILs have written language. Both languages used in this work have established standards of writing. 


\emph{Guarani Mbya} and \emph{Nheengatu} are two of such Indigenous languages of Brazil, belonging to the same \textit{Tupi-Guarani} family of the \textit{Tupi} branch. However, they have diverged significantly in their development, reflecting the adaptability of Indigenous languages to historical pressures and to quite distinct evolutionary paths. 

\emph{Guarani Mbya}, spoken by Indigenous communities in Brazil, Paraguay, and Argentina, has largely retained its traditional structures. It is an agglutinative, head-marking language with flexible word order, typically favoring a subject-verb-object (SVO) structure but there is a substantial presence of the subject-object-verb (SOV) structure, specially in subordinate clauses. The language's morphology is complex, with verbs cross-referencing their arguments through distinct affixes for active and inactive verbs. This complexity underscores the language's strong ties to its Tupi-Guarani roots, even as it has adapted to the influence of dominant languages like Portuguese and Spanish~\citep{kiss-thomas-2019-word,dooley2016}. Different forms of the Guarani language have had a written form since the arrival of the Portuguese in the 1500s, and one of its major variants, the \textit{Paraguayan Guarani} is spoken by more than 5~million people today.

\emph{Nheengatu} evolved from Tupinambá, a dialect spoken by coastal Indigenous groups. However, its historical development was shaped by significant contact with Portuguese during Brazil's colonial period, especially as Jesuit missionaries promoted it as the \textit{lingua franca} of the Amazon region. This interaction led to extensive borrowing from Portuguese and simplifications in its grammar, making Nheengatu more accessible to a broader population. Despite these changes, Nheengatu remains an agglutinated language, like Guarani Mbya, and retains the SVO word order~\citep{moore1994nheengatu}. However, its morphology is less complex, reflecting its role as a tool for wider communication rather than a strictly preserved Indigenous language~\citep{navarro2016}. Moreover, it is a language that has been adopted by several Indigenous communities after the loss of their native tongue, specially in the Alto Rio Negro region~\citep{storto2019}, by communities such as the \textit{Bar\'e}. The language has a strong community of writers and translators, who did the first translation of the Brazilian constitution to an Indigenous language~\citep{constitution_nheengatu_2023}.

While both languages share a foundation in the Tupi-Guarani family, their morphological and syntactic differences are pronounced. We believe that such differences may present challenges in terms of creating MT models, as explored next.


\section{Datasets, Models, and Metrics}


This section describes the two datasets used in this study, the models employed as the base of the finetuning process, and the evaluation metrics used. The raw datasets were obtained from the authors of~\citep{pinhanez2024harnessing}

\subsection{The Guarani Mbya Dataset}

Sentences from three different sources were used in the construction of Guarani Mbya dataset according to~\citep{pinhanez2024harnessing}. The first source was a set of Guarani Mbya short stories with 1,022 sentences, also available in Portuguese and English~\citep{dooley1988a,dooley1988b}. The second comprises 245 texts extracted from PDF files with a pedagogical character~\citep{dooley1985}. The third source was Robert A. Dooley’s Lexical Guarani Mbya dictionary~\citep{dooley2016}, a reference work for the language, from which we extracted 2,230 sentence pairs. The last two sources contained sentence pairs in Guarani Mbya and Portuguese only. The data from the three sources was concatenated and cleaned, removing non-alphanumeric characters and normalizing Unicode values. 

We performed further cleaning by removing duplicates, ending with a dataset of 3440 paired sentences with translations from Guarani Mbya to Portuguese. To ensure the presence of all sources in training and testing, we divided each source in train and test, concatenating all trains in the final train with 3319 sentences pairs and all tests in a test set with 121 sentences pairs. 
In Table~\ref{tab:gun_dataset_stats} we present some statistics about the Guarani Mbya dataset.

\subsection{The Nheengatu Dataset}

According to~\citep{pinhanez2024harnessing}, the original dataset for Nheengatu consists of texts extracted from several documents with translations to Portuguese. Such documents inclued varied content, such as a novel~\citep{ishikawa_brilhos_2019}, teaching material~\citep{navarro2016}, and a lexicon~\citep{avila_2021}. Such documents were processed by~\citet{pinhanez2024harnessing}, both in automated and manual procedures, and converted to a set of 7180 paired sentences. 

To prepare further the received data, we conducted a data cleaning process with the following steps. Initially, 226 sentences (3.1\% of original data) were removed because of duplication. We removed also documents that included comments, observations and literal translations in parentheses or brackets in the Portuguese translation. Since there is not the ``to be'' verb in Nheengatu, but some documents included it in parentheses in the Portuguese translation to highlight the absence of it in the original language, we conducted a normalization step to avoid random parenthesis in translation and we removed the parentheses and kept the text, including the verb, in the sample.

After those normalizations, we removed another batch of repeated or duplicated samples in either the source or target side of the pair, since a new set of duplications emerged. This process resulted in a total of 6932 sentences pair with translations from Nheengatu to Portuguese. We then randomly divided this dataset into 6699 sentence pairs for training, and the remaining 233 sentence pairs comprise the test set. In Table~\ref{tab:yrl_dataset_stats} we present some statistics about the Nheengatu dataset.

\subsection{The Pre-Trained NLLB Models}
Given that NLLB has demonstrated promissing results with Indigenous and low-resource languages, we selected it as the foundation for fine-tuning. 

\begin{table}[t!]

\caption{The Guarani Mbya dataset.}
\label{tab:gun_dataset_stats}

\begin{tabular}{lrr}
\hline
\textbf{Guarani Mbya}    & \textbf{\begin{tabular}[c]{@{}c@{}} training \end{tabular}} & \textbf{\begin{tabular}[c]{@{}c@{}} testing \end{tabular}} \\ \hline
\begin{tabular}[c]{@{}c@{}}vocabulary size\end{tabular}                     & 4229                                                                  & 430                                                                  \\ \hline
\begin{tabular}[c]{@{}c@{}}words per sentence\end{tabular}                  & 1.27                                                                  & 3.55                                                                 \\ \hline
\begin{tabular}[c]{@{}c@{}}minimum length\end{tabular}                      & 1                                                                     & 2                                                                    \\ \hline
\begin{tabular}[c]{@{}c@{}}maximum length\end{tabular}                      & 82                                                                    & 40                                                                   \\ \hline
\begin{tabular}[c]{@{}c@{}}mean sentence length\end{tabular}               & 6.78                                                                  & 6.49                                                                 \\ \hline
\begin{tabular}[c]{@{}c@{}}std dev  sentence length\end{tabular} & 5.42                                                                  & 4.87                                                                 \\ \hline
\end{tabular}
\end{table}

Notice that the 200~languages of NLLB have millions of speakers, and although the languages have considerably lower amounts of data available in comparison to English and Chinese, they are in a completely different level of data availability than ultra-low resource languages such as Guarani Mbya or Nheengatu.

In this work we used two variations of NLLB-200: the NLLB-600M\footnote{https://huggingface.co/facebook/nllb-200-distilled-600M} with 600 million parameters and NLLB-3B\footnote{https://huggingface.co/facebook/nllb-200-3.3B} with 3.3 billion parameters. As a tokenizer, we instructed the model to use the one for the Paraguayan Guarani language, one of the 200~training languages. Both models were finetuned for 57,000 training steps, with a learning rate learning rate schedule that linearly increases from zero over the first 1000 steps and then remains constant at $10^{-4}.$ The model was trained with batch size of 16 and a maximum lenght of 128 tokens, producing favorable results for Nheengatu, so they were not subject to further optimization, and all models were  trained with this configuration. 

\begin{table}[t!]

\caption{The Nheengatu dataset.}
\label{tab:yrl_dataset_stats}

\begin{tabular}{lrr}
\hline
\textbf{Nheengatu}    & \textbf{\begin{tabular}[c]{@{}c@{}} training\end{tabular}} & \textbf{\begin{tabular}[c]{@{}c@{}} testing\end{tabular}} \\ \hline
\begin{tabular}[c]{@{}c@{}}vocabulary size\end{tabular}                      & 9159                                                               & 635                                                               \\ \hline
\begin{tabular}[c]{@{}c@{}}words per sentence\end{tabular}                  & 1.37                                                               & 2.73                                                              \\ \hline
\begin{tabular}[c]{@{}c@{}}minimum length\end{tabular}                      & 1                                                                  & 1                                                                 \\ \hline
\begin{tabular}[c]{@{}c@{}}maximum length\end{tabular}                    & 38                                                                 & 26                                                                \\ \hline
\begin{tabular}[c]{@{}c@{}}mean sentence length\end{tabular}                & 7.18                                                               & 4.63                                                              \\ \hline
\begin{tabular}[c]{@{}c@{}}std dev sentence length\end{tabular} & 4.77                                                               & 3.39                                                              \\ \hline
\end{tabular}
\end{table}

\subsection{Evaluation Metric}
For the automatic evaluation of the models, we rely on the chrF metric \citep{popovic-2015-chrf}, a commonly-used choice for low-resource languages \citep{ebrahimi-etal-2024-findings}. We consider segment-level aggregation \citep{cavalin2024sentencelevelaggregationlexicalmetrics}, using the \verb|sentence_chrf| function from SacreBLEU\footnote{https://github.com/mjpost/sacrebleu}.

\section{Experiments and Results}
In this section we present the experiments we performed and their results. We used the cleaned versions of the training datasets, processed from the data we got from~\citep{pinhanez2024harnessing} and, unlike in that work, covered both directions of translation. We first describe a zeroshot evaluation of the NLLB-600M model, followed by a basic finetuning process with the two full training datasets. We then explore the impacts of using pre-trained models of different sizes and, following, different sizes of the Nheengatu training dataset. A summary with the mean chrF scores and respective standard deviations of all experiments is presented in Table~\ref{tab:results_summary} and the respective distributions are provided in detail in Table~\ref{tab:results_full} of the Appendix~\ref{app:detailed_experiment_results}.

\subsection{Zeroshot Evaluation}
Given that NLLB models have been trained with Paraguayn Guarani as one of the 200 pre-training languages, which is a closely-related language to Guarani Mbya and also belongs to the Tupi-Guarani family, such as both Mbya and Nheengatu, we first conducted a zeroshot evaluation with the NLLB-600M model to evaluate the quality of these models before any finetuning with our own data.

\begin{figure}[t!]
    \centering
    \begin{subfigure}{0.47\textwidth}
        \centering
        \includegraphics[width=0.9\textwidth]{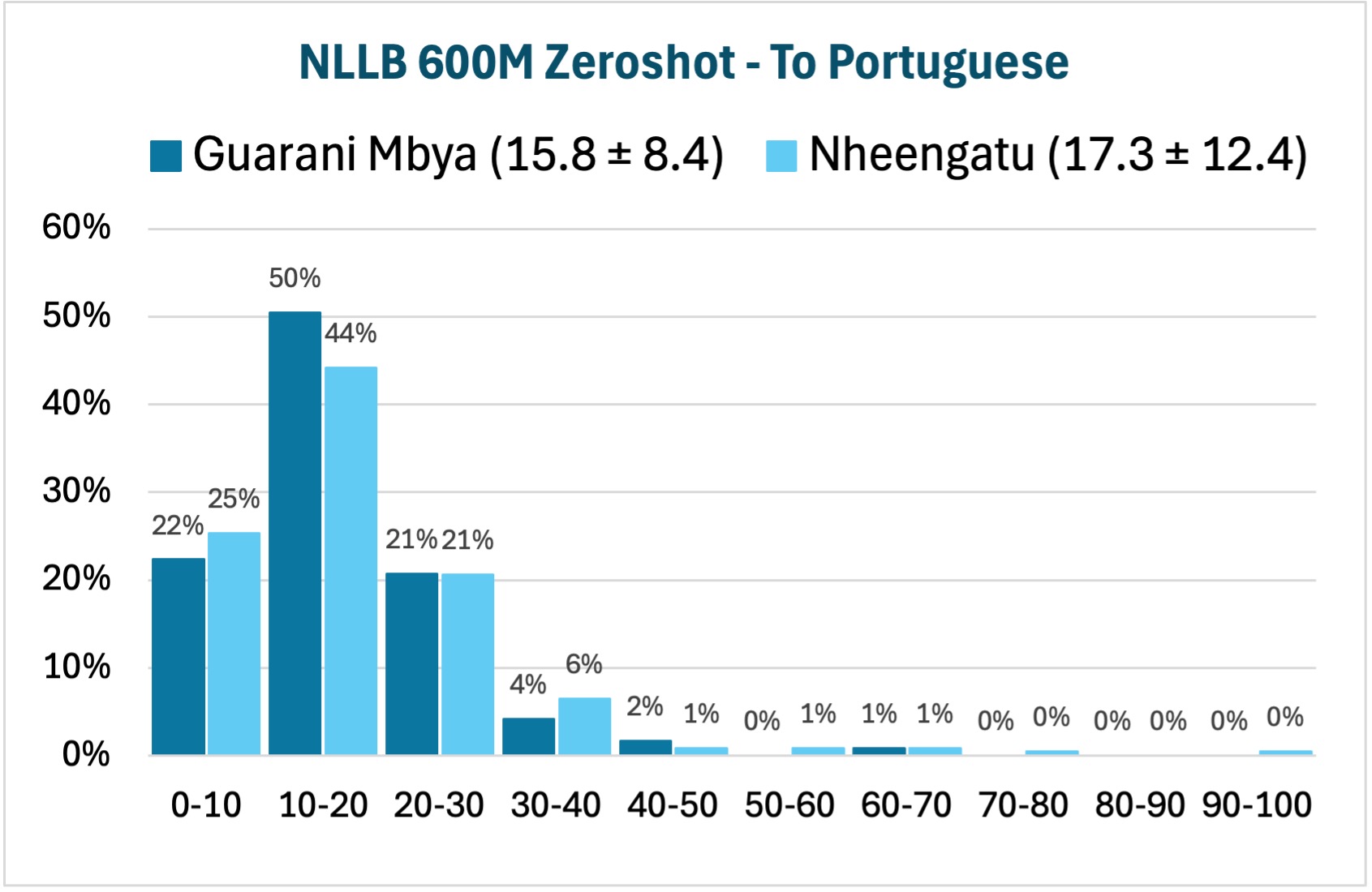}
        \caption{To Portuguese.}
        \label{fig:zeroshot_xx_por}
    \end{subfigure}
    \\ \vspace{2mm}
    \begin{subfigure}{0.47\textwidth}
        \centering
        \includegraphics[width=0.9\textwidth]{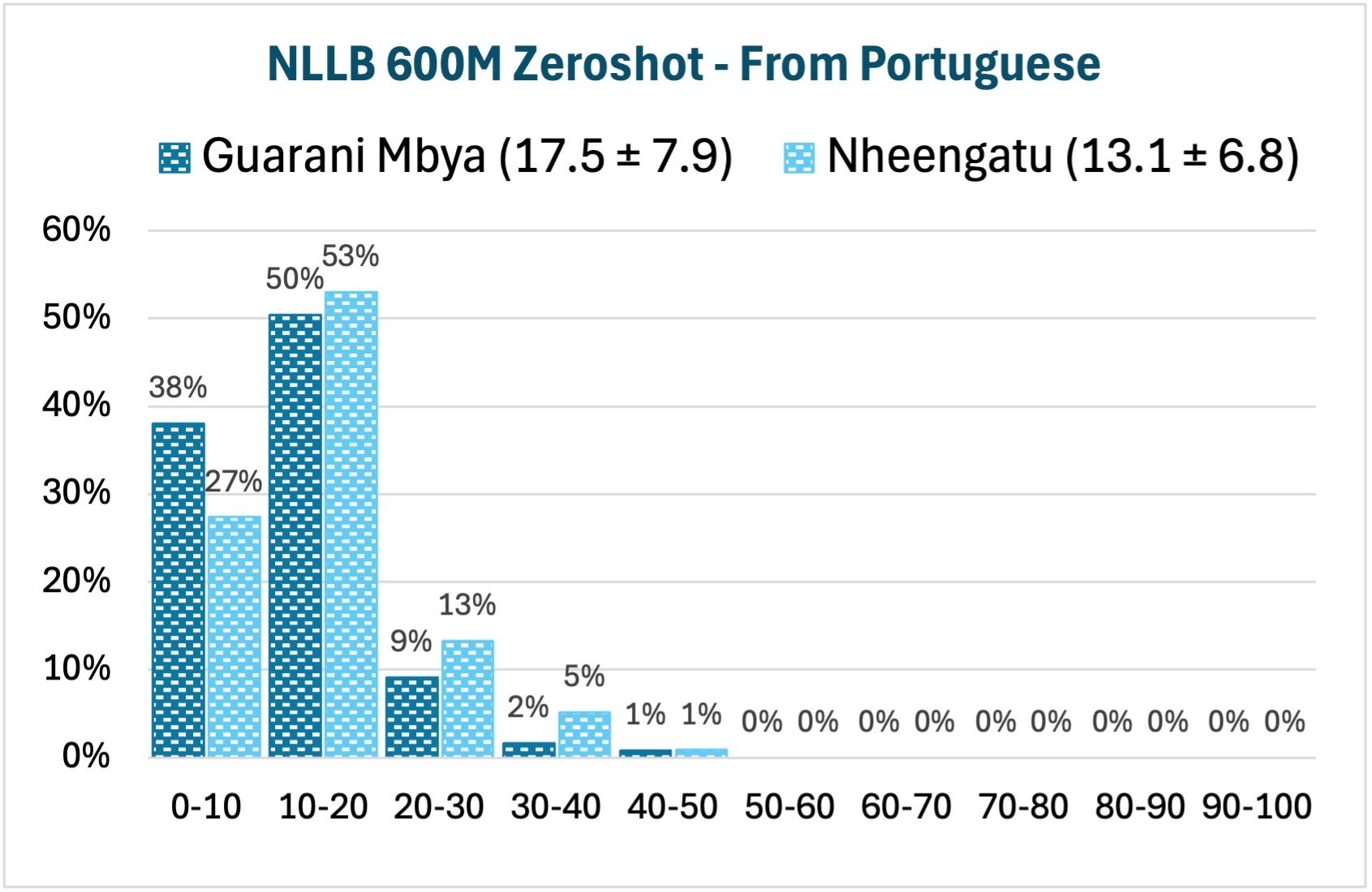}
        \caption{From Portuguese.}
        \label{fig:zeroshot_por_xx}
    \end{subfigure}
    \caption{Distribution of the evaluation of the Zeroshot NLLB-600M model for Guarani Mbya and Nheengatu.}
    \label{fig:zeroshot_evaluation_nllb600}
\end{figure}

In Figure~\ref{fig:zeroshot_evaluation_nllb600}.a we present the distribution of scores with for translating the test set between both Indigenous languages and Portuguese. We can observe relatively similar distributions, but on average we observe slightly worse results for inputs in Guarani Mbya 15.8 ($\sigma$: 8.4, where $\sigma$ denotes the standard deviation) while for Nheengatu the mean chrF observed is of 17.3 ($\sigma$: 12.4).

\begin{figure}[t!]
    \centering
    \begin{subfigure}{0.45\textwidth}
        \centering
        \includegraphics[width=0.9\textwidth]{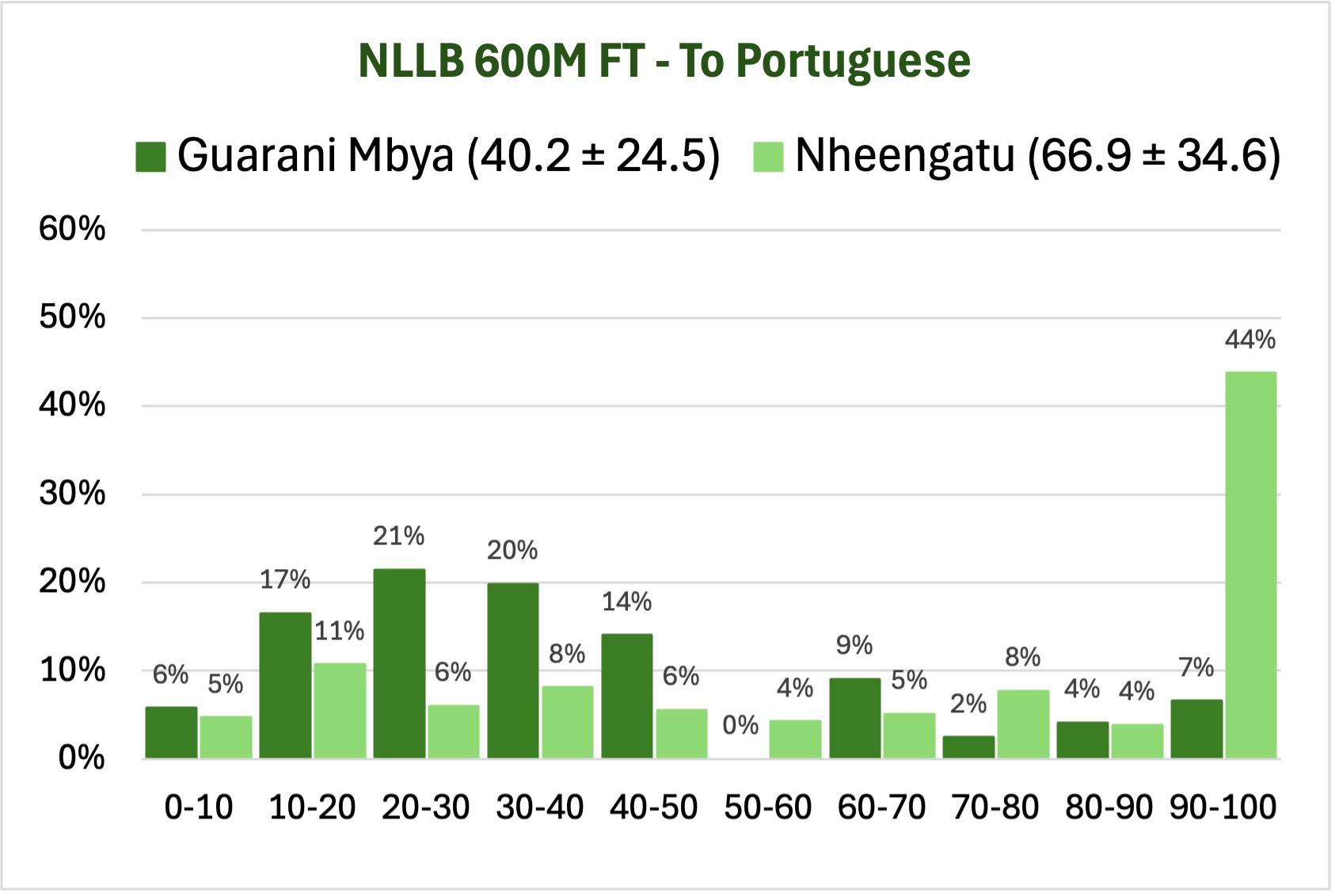}
        \caption{To Portuguese.}
        \label{fig:comparison_mbya}
    \end{subfigure}
    \\ \vspace{2mm}
    \begin{subfigure}{0.45\textwidth}
        \centering
        \includegraphics[width=0.9\textwidth]{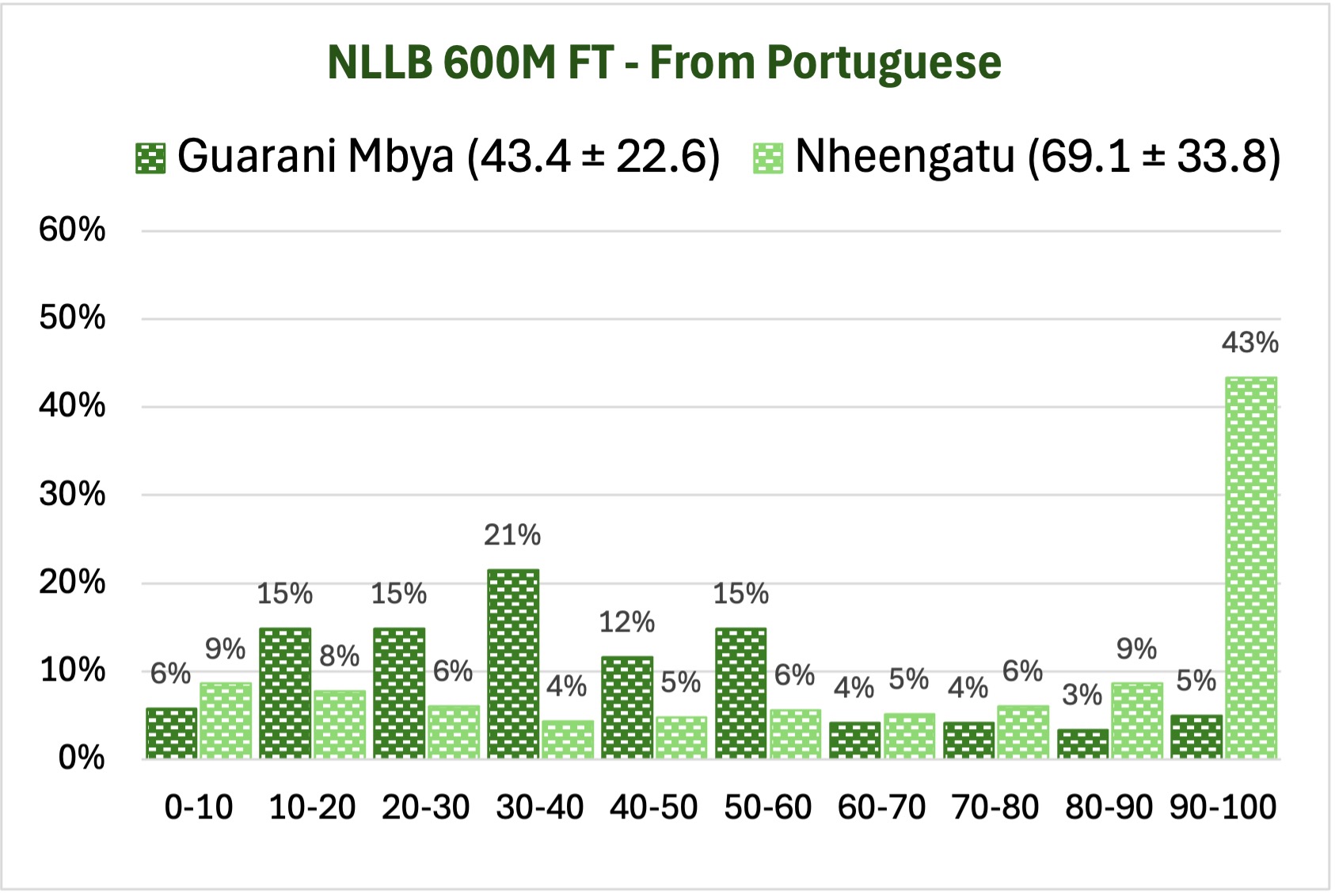}
        \caption{From Portuguese.}
        \label{fig:comparison_nheengatu}
    \end{subfigure}
    \caption{Distribution of evaluation of the finetuned NLLB 600M models for Guarani Mbya and Nheengatu.}
    \label{fig:fine_tuning_NLLB600}
\end{figure}



On the other hand, when the translation is in the opposite direction, the mean chrF scores for Guanari Mbya is slightly higher, as shown in Figure~\ref{fig:zeroshot_evaluation_nllb600}.b.  In the Indigenous languages to Portuguese translation directions, the mean chrF score is of 17.5 ($\sigma$: 7.9) for Guarani Mbya, while for Nheengatu the mean chrF observed is of 13.1 ($\sigma$: 6.8). 

The results clearly show that the zeroshot use of NLLB-600M for translation of both languages is not a feasible option. More important for our analysis, the performance of both translators is basically identical for both languages, both considering the mean values and the distributions.

\subsection{Basic Finetuning Evaluation}
The following experiments focus on comparing the impact after the models are finetuned with the standard training datasets. 
In Figure~\ref{fig:fine_tuning_NLLB600} we present the comparison of the distribution of the values for the Guarani Mbya and Nheengatu, in both translation directions.

Considering the Guarani Mbya to Portuguese translation direction, we observed a mean chrF score of 40.2 ($\sigma$: 24.6), while the Nheengatu mean score was considerably higher, of 66.9 ($\sigma$: 34.6). Observing Figure~\ref{fig:fine_tuning_NLLB600}.a, it is clear to see that the main difference is the high number of near-perfect scores of the latter. Figure~\ref{fig:fine_tuning_NLLB600}.b shows the opposite direction of translation, with very similar results: the Guarani Mbya translator had a mean score of 43.4 ($\sigma$: 22.6) while the Nheengatu translator had a score of 69.1 ($\sigma$: 33.8), with very similar distribution profiles for both translation directions.

Moreover, when compared to the zeroshot performance picture in Figure~\ref{fig:zeroshot_evaluation_nllb600}, we see stark differences between the translators of the two languages. While the Guarani Mbya translators, in both directions, clearly improved in relation to the zeroshot versions, the performance of the Nheengatu translators is significantly better than both the zeroshot and the Guarani Mbya finetuned versions. These results are very similar to the ones reported in~\citep{pinhanez2024harnessing} which were obtained with the WMT-19 pre-trained model. Further, our work indicates that those differences happen in both directions of translation. There are strong differences in performance between the two languages and the following experiments explored two factors which could possibly explain the differences: the size of the pre-trained models and the training dataset sizes.

\begin{figure}[t!]
    \centering
    \begin{subfigure}{0.45\textwidth}
        \centering
        \includegraphics[width=0.9\textwidth]{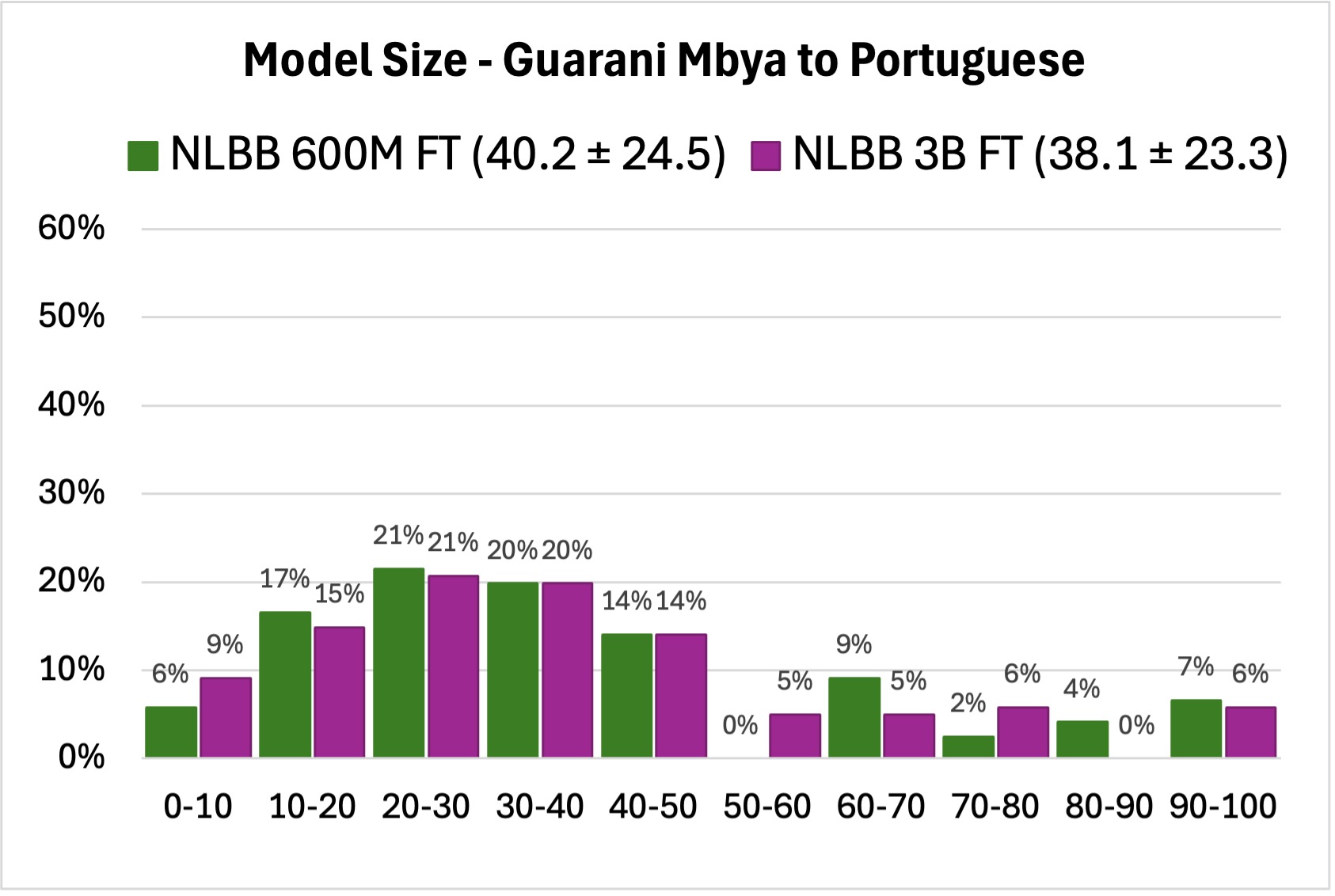}
        \caption{To Portuguese.}
        \label{fig:ModelSize_gun_por}
    \end{subfigure}
    \\ \vspace{2mm}
    \begin{subfigure}{0.45\textwidth}
        \centering
        \includegraphics[width=0.9\textwidth]{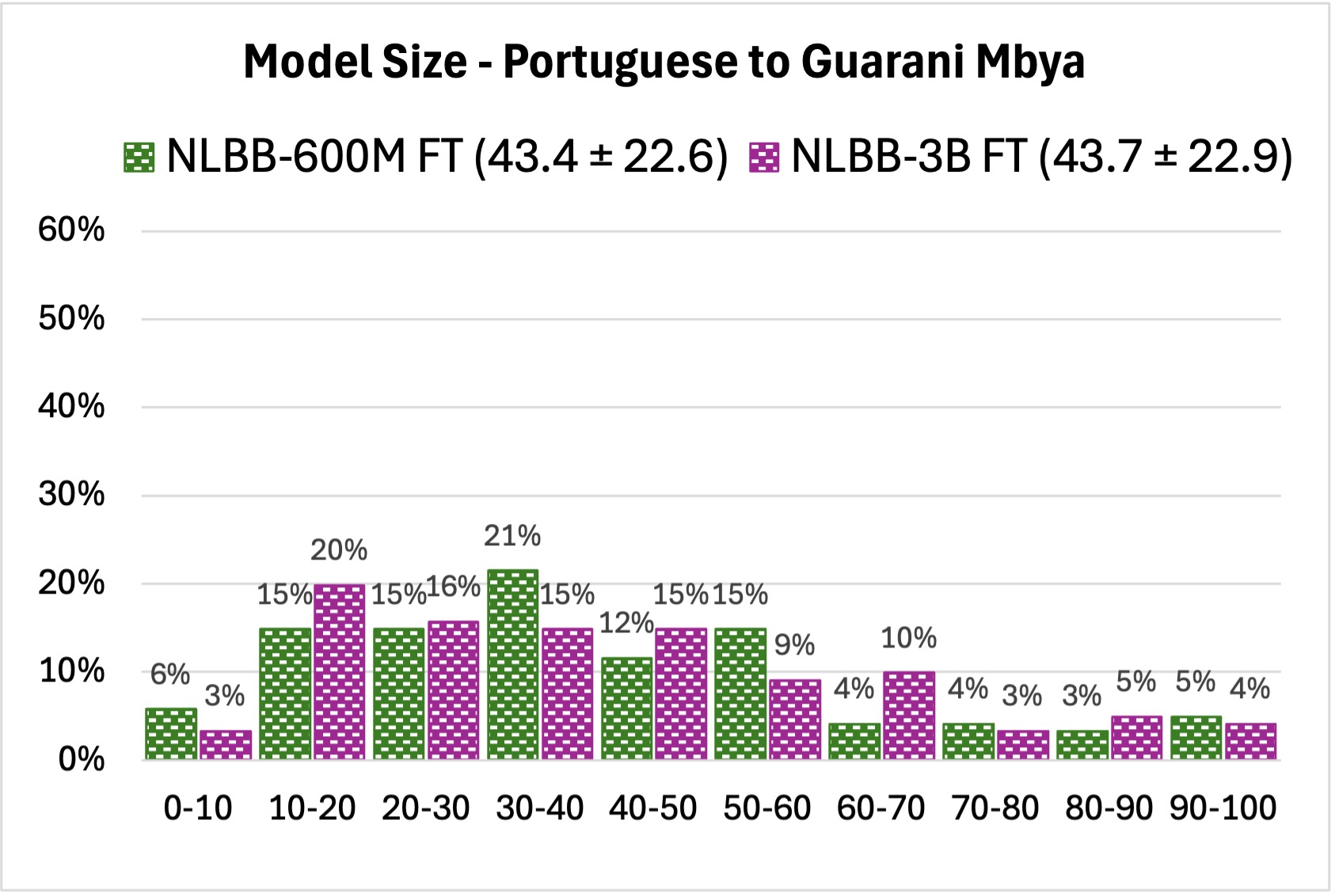}
        \caption{From Portuguese.}
        \label{fig:ModelSize_por_gun}
    \end{subfigure}
    \caption{Evaluation of different sizes of NLBB model (600M and 3B parameters) for Guarani Mbya.}
    \label{fig:ModelSize_Guarani}
\end{figure}

\begin{figure}[t!]
    \begin{subfigure}{0.45\textwidth}
        \centering
        \includegraphics[width=0.9\textwidth]{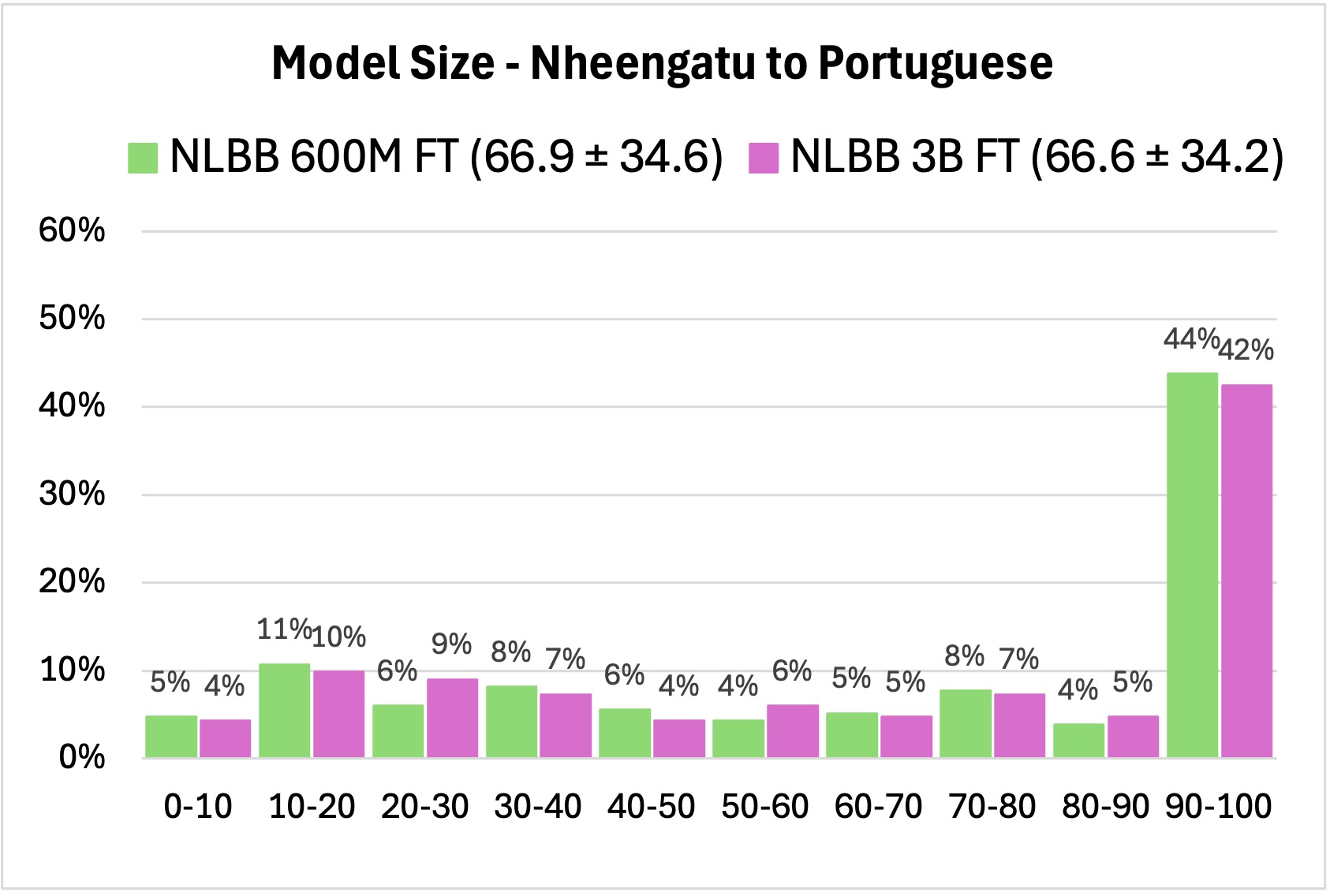}
        \caption{To Portuguese.}
        \label{fig:ModelSize_yrl_por}
    \end{subfigure}
    \\ \vspace{2mm}
    \begin{subfigure}{0.45\textwidth}
        \centering
        \includegraphics[width=0.9\textwidth]{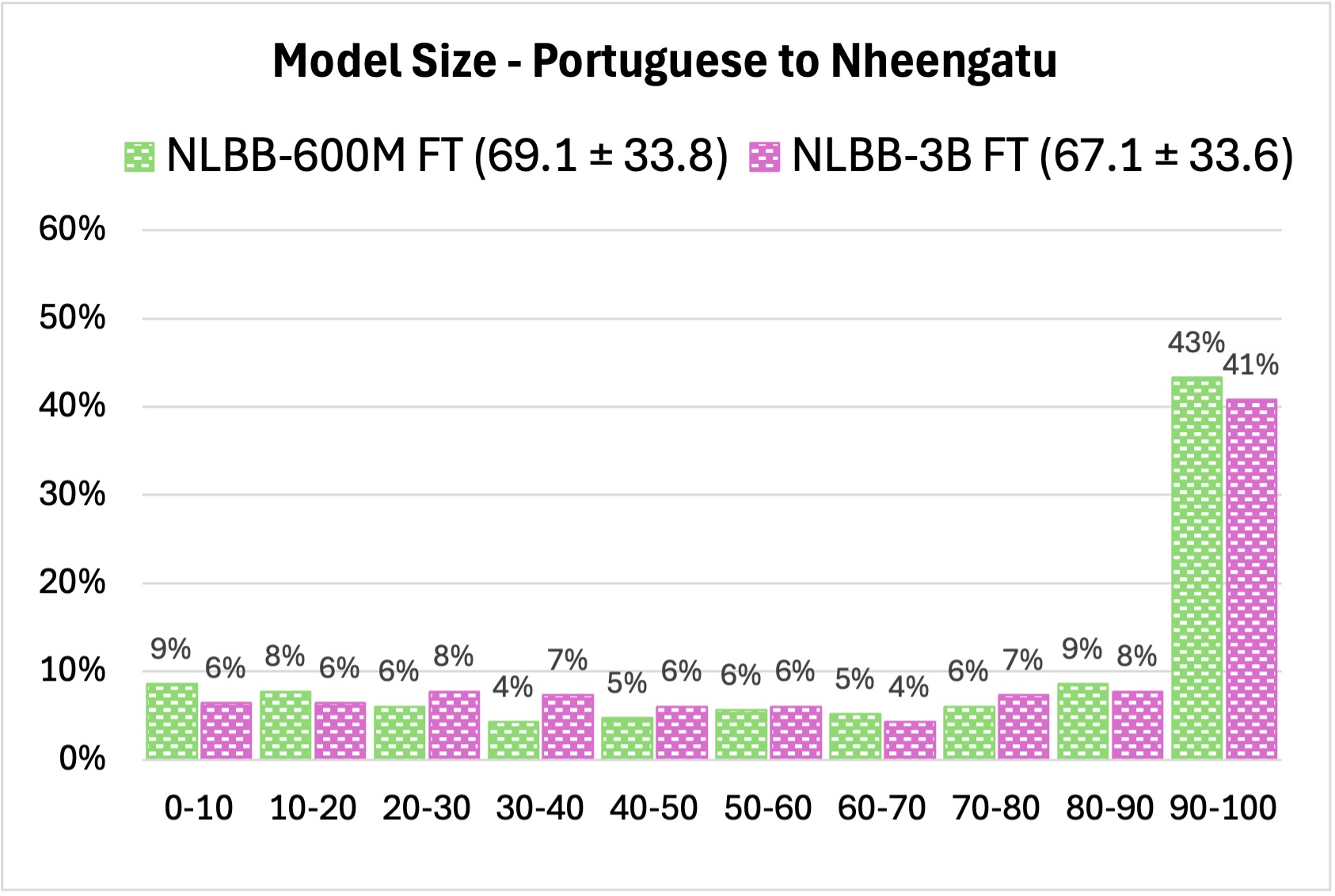}
        \caption{From Portuguese.}
        \label{fig:ModelSize_por_yrl}
    \end{subfigure}
    \caption{Evaluation of different sizes of NLBB model (600M and 3B parameters) for Nheengatu.}
    \label{fig:ModelSize_Nheengatu}
\end{figure}

\begin{figure*}[t!]
    \centering
    \begin{subfigure}{\textwidth}
        \centering
        \includegraphics[width=\textwidth]{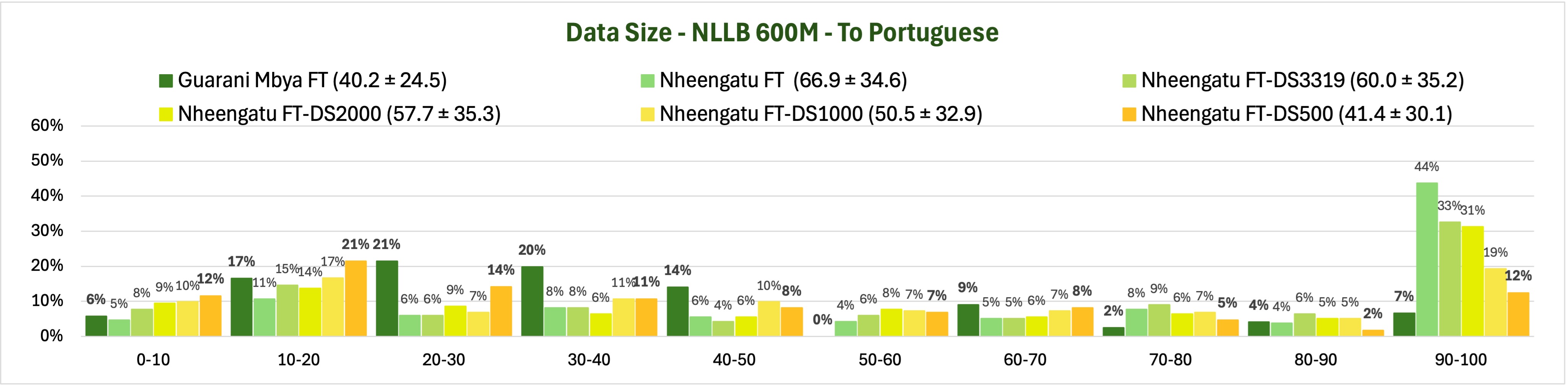}
        \caption{To Portuguese.}
        \label{fig:downsampling_yrl-por}
    \end{subfigure}
    \\ \vspace{2mm}
    \begin{subfigure}{\textwidth}
        \centering
        \includegraphics[width=\textwidth]{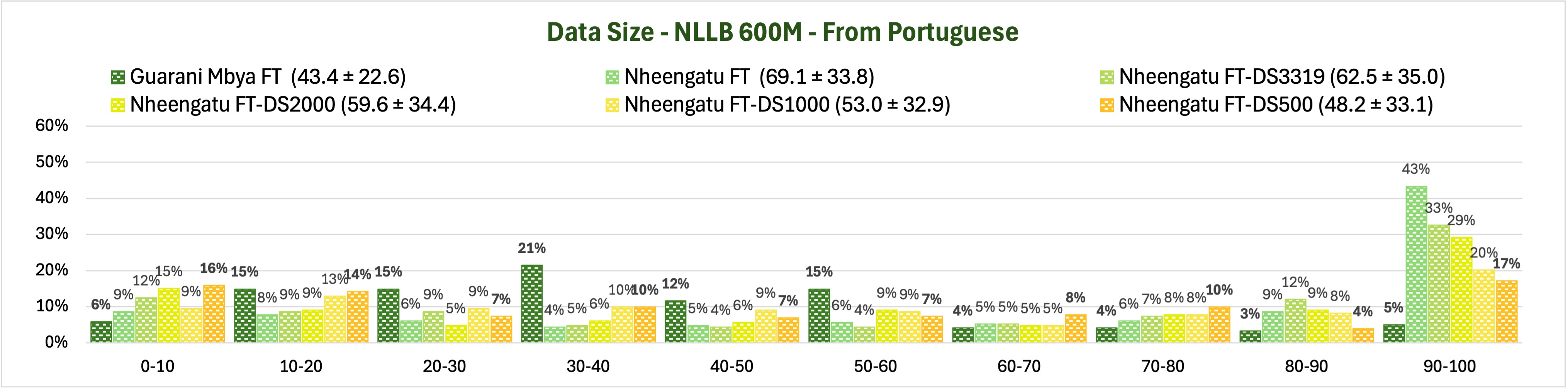}
        \caption{From Portuguese.}
        \label{fig:downsampling_por-yrl}
    \end{subfigure}
    \caption{Comparison of downsampling strategies for the Nheengatu training dataset and the Guarani Mbya baseline.}
    \label{fig:nheengatu_downsampling}
\end{figure*}

\subsection{Model Size Impact Evaluation}
Although similar results were obtained with the WMT-19 pre-trained model~\citep{pinhanez2024harnessing} and the NLLB-600M model, we wanted to check whether the observed differences between the Guarani Mbya and Nheengatu translators were not caused by limitations in the model sizes, which are, in fact, similar in terms of number of parameters.

To explore this, we considered two versions of NLLB with different sizes, i.e. NLLB-600M and NLLB-3B, which we used as the base versions for the finetuning process. Considering the Guarani Mbya language, Figure~\ref{fig:ModelSize_Nheengatu} shows the distributions of results for both models. The distributions are almost identical, for both directions of translation, and the mean scores are 40.2 ($\sigma$: 24.5) and 38.2 ($\sigma$: 23.3) for the Guarani Mbya translators to Portuguese finetuned using NLLB-600M and NLLB-3M, respectively. In the opposite translation direction, the results were also very similar, with mean scores are 43.3 ($\sigma$: 22.6) and 43.7 ($\sigma$: 22.9) for the Portuguese to Guarani Mbya translators finetuned using NLLB-600M and NLLB-3M, respectively.

The results for the Nheengatu translators were strikingly similar, as shown in Figure~\ref{fig:ModelSize_Nheengatu}.
Considering Neengatu-to-Portuguese translations, with NLLB-600M a score of 66.9 ($\sigma$: 34.6) was achieved and an score of 66.6 ($\sigma$: 34.2) with NLLB-3B; in the Portuguese to Nheengatu direction, a mean score of 69.11 ($\sigma$: 33.8) was achieved with NLLB-600M, and 67.1 ($\sigma$: 33.6) with NLLB-3B. The distributions look basically similar.

These experiments seem to indicate that the size of the model has no impact on the difference of performance between the Guarani Mbya and the Nheengatu translator, for both directions. Combined with the previous results reported in~\citep{pinhanez2024harnessing}, it seems to point towards causes which are not related to the pre-trained model.

\subsection{Training Dataset Size Impact Evaluation}

Another possibility to explain the gap between the results presented so far is that the size of the training sets are quite different, i.e. the training set for Guarani Mbya, 3319~pairs, is about half of the size of the Nheengatu training set, 6699 pairs. 

We first explored this by making both training sets to have the same size, by downsampling the Nheengatu training dataset to have only 3319~pairs, resulting in a dataset that we refer to as DS3319. As shown in Figure~\ref{fig:nheengatu_downsampling}, there was some degradation in performance, from 66.9 ($\sigma$: 34.6) to 60.0 ($\sigma$: 35.2) in the translation to Portuguese, and from 69.1 ($\sigma$: 33.8) to 62.5 ($\sigma$: 35.0) in the opposite direction. Notice that the gaps between Nheegantu and Guarani Mbya were originally of 26.7 and 25.7 score points, and they only decreased to 19.8 and 19.1 with this downsample. 

\begin{table*}[t!]
    \centering
    \caption{Summary with the mean chrF scores and standard deviations of the experiments.}
    \label{tab:results_summary}
    \begin{tabular}{llcccc}
    \toprule
    Method & Base Model & Guarani Mbya & Nheengatu & Portuguese to & Portuguese to \\
    &       & to Portuguese & to Portuguese & Guarani Mbya & Nheengatu \\
    \midrule
     Zeroshot    &  NLLB-600M &    15.8 (8.4) &   17.3 (12.4) &   17.5 (7.9)  &  13.1 (6.8) \\
     FT          &  NLLB-600M &    \textbf{40.2 (24.5)} &   \textbf{66.9 (34.6)} &   \textbf{43.4 (22.6)}  &  \textbf{69.1 (33.8)} \\
     FT          &  NLLB-3B   &    38.1 (23.3) &   \textbf{66.6 (34.2)} &   \textbf{43.7 (22.9)}  &  67.1 (33.6) \\
     FT-DS3319   &  NLLB-600M &     -          &   60.0 (35.2) &   -            &  62.5 (35.0) \\
     FT-DS2000   &  NLLB-600M &     -          &   57.7 (35.3) &   -            &  59.6 (34.4) \\
     FT-DS1000   &  NLLB-600M &     -          &   50.5 (32.9) &   -            &  53.0 (32.9) \\
     FT-DS500    &  NLLB-600M &     -          &   41.4 (30.1) &   -            &  48.2 (33.1) \\
    \bottomrule
    \end{tabular}

\end{table*}

We then investigated whether there was a reduction in training data size that would make the performance for Nheengatu comparable to the poor performance of the Guarani Mbya translators. For that, we created three downsampled datasets named DS2000, DS1000, and DS500, which contain, respectively, 2,000, 1,000, and 500 pairs. 

The results with all downsampled datasets are also depicted in Figure~\ref{fig:nheengatu_downsampling}.
We can observe that we were only able to degrade the performance to Nheengatu to be comparable to Guarani Mbya, when the downsampled training set reached only 500 samples, about 1/7 of the training data of Nheengatu. In details, always considering first the Nheengatu-to-Portuguese direction and then the opposite one, we can observe the following results: with 2,000 samples, there is a reduction of 9.2 and 7.7 in the score; with 1,000, the reduction is of 16.4 and 16.1; and with 500 samples, 25.5 and 20.9.

Our results convincingly show that having a twice-as-large training dataset does not seem to be the reason for the gap between the Nheengatu and Guarani Mbya translators. 
Moreover, we conducted a polynomial regression with degree 4 with all five Nheengatu datasets using their chrF mean scores in the direction Nheengatu to Portuguese, to relate the dataset size and the model performance. 
The first regression model estimated the relationship between the number of sentences and the chrF score, estimating that 369 sentences in Nheengatu were necessary to achieve the same 38.1 chrF score achieved by Guarani Mbya. 
Another way to see this result is to suppose a linear relation between the number of sentences and the chrF metric (1 to 7). In this scenario it would be necessary to have 60,255 pairs in the  Guarani Mbya training set to achieve a similar performance then the Nheengatu translators. In our view, the results show that the size of the training set is not sufficient to explain the differences in performance of the  translators.




\subsection{Discussion}

Table~\ref{tab:results_summary} presents the mean chrF score and the standard deviations of all the explored translators. It shows that the best results were obtained with the finetuning with the standard training datasets of the smaller model NLLB-600M, although the high standard deviations possibly signal that those differences are not too significative. It also shows that zeroshoting is not an option and that the direction of the translation yields similar results.

More importantly for our analysis, the results indicate that the size of the pre-trained model does not make a difference, and that there is limited impact from the training data size. Of course there could be other data-related factors affecting the results, such as qualitative aspects of the training data used for the Guarani Mbya and the Nheengatu translators, such as sentence complexity, nature of vocabulary, or language consistency. 

However, we believe that our experiments point towards the possibility that structural linguistic differences between the languages, such as the predominance of SOV or SVO structures, may play an important role in the final performance of the finetuned translators. If so, it may be required that new translation techniques or technologies must be developed to handle certain classes of ultra-low resource languages such as the one which Guarani Mbya belongs. If this is true, not only it opens up rich research areas of finding distinctive language features which impact finetuning and of developing their associated technologies but it creates opportunities to better understand how finetuning and model-based translation actually work.

\section{Final discussion}
In this work we presented an investigation of the possible causes of the performance differences between the finetuning of a pre-trained model for Guarani Mbya and Nheengatu, a pair of related Brazilian Indigenous languages. We showed that finetuning does improve the performance of the translators in both directions of the two languages, compared with zeroshot evaluations, but the improvements are much more salient for Nheengatu. 

We also provided evidence that the model size does not affect the performance of the translators. We also showed that the size of the training size influences the performance but is not able to explain it completely.
We think it is hard to believe that the performance differences are just a matter of training data but an indication that the now-traditional method of finetuning pre-trained models may not apply to all languages. As we mentioned in Section~\ref{bils}, Nheengatu is majorly an SVO language owing to its influence from Portuguese, while Guarani Mbya presents a mix of SVO and SOV language structure. Besides that, the latter also is considerably more complex in morphology. Such differences make us question whether structural and morphological complexity of languages like Guarani Mbya may not only have a negative impact in the finetuning 
but also whether the current finetuning strategy is effective for such types of languages.

Different research directions may allow us to explore further those issues. 
First, we want to perform studies in other languages with different structural linguistic characteristics, to provide a better range of explaining hypothesis.
Another opportunity is to create translators for languages which share similar features to the set of languages used to create pre-trained models. For that, we could either focus on a model created from scratch or on trying to incorporate such languages to an existing model, for instance by conducting an intermediate incremental learning step to incorporate new pre-training languages. Another research direction is to train with synthetic data which emulate the characteristics of an ultra-low resource language. For example, one could create a synthetic language based on obfuscating an existing one \citep{he-etal-2023-synthetic}.

\section{Limitations}
One main limitation of this work is that we rely solely on automated metrics, while qualitative analysis could a valuable complement to help elucidate the differences in results between Nheengatu and Guarani Mbya. For this work we rely on the evidence provided in \citep{pinhanez2024harnessing}, attesting the superior quality of Nheengatu translators created from finetuning the WMT19 pre-trained model, but we have not conducted a similar evaluation for the translator built on top of NLLB-200.

Another limitation is the work relies only on two languages and we think it is highly desirable to expand our analysis to more languages. We have plan to work with other Brazilian Indigenous languages in the near future, but given the difficulties in collecting data for this kind of language, we were not able to include such analysis at this moment.

Finally, we believe that a more fine-grained evaluation of the results also can be beneficial to better understanding of outcomes from this research. Since Guarani Mbya can present both SOV and SVO sentences, measuring translation quality in subsets of sentences that share similar structure could provide us a clearer picture if the structure of the language is something that is actually impacting the results. Unfortunately, to have such subsets we would need to carry out an annotation process which was not possible at this time.

\section{Ethical Statement}
While conducting this research, we were mindful of the ethical considerations involved in work with Indigenous communities. Additionally, we recognize that a significant aspect of colonial history, both past and present, has been the spread of various forms of Christianity among Indigenous peoples. Consequently, the Bible is frequently one of the most available texts translated into these languages. The use of religious data, especially the Bible dataset is dangerous and its cultural toxicity is well described in \citep{domingues_etal2024_quantifying}. To avoid introducing religious bias or potentially toxic results, we deliberately chose not to include Bible or other religious datasets in our training data.

Furthermore, all authors whose works contributed to our dataset were informed of and consented to the use of their materials for this study.

Moreover, we uderstand the complex political and ideological aspects of working with language vitalization. To enhance the conduct of our research, we follow some protocols proposed in \citep{pinhanezetal2023balancing}.

\bibliography{custom}

\clearpage

\noindent\textbf{\large Appendixes}

\appendix

\section{Detailed Experiment Results}
\label{app:detailed_experiment_results}

Table~\ref{tab:results_full} displays in detail the experimental results, showing the number of testset pairs which produced chrF scores for each bin of the distributions
and their associated percentages.

\begin{table*}[h!]
    \centering
    \includegraphics[width=\textwidth]{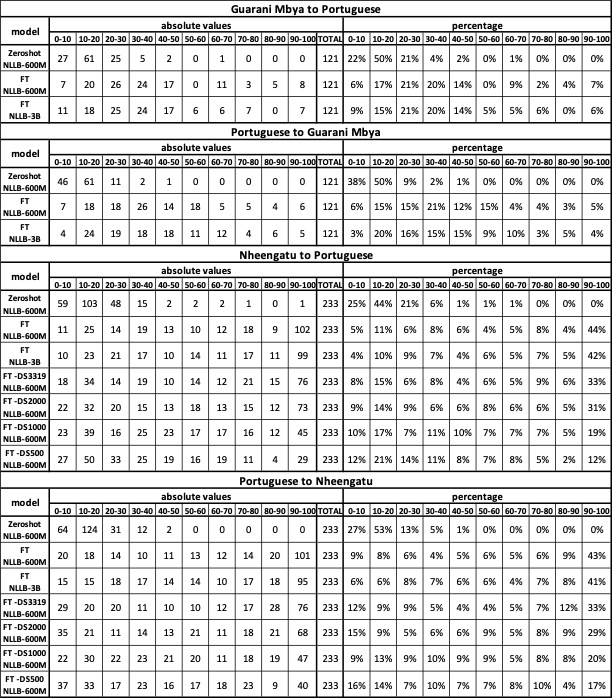} 
    \caption{Full experimental results. These results are summarized in Table~\ref{tab:results_summary}}
    \label{tab:results_full}
\end{table*}

\end{document}